\documentclass{article}

\usepackage{PRIMEarxiv}

\usepackage[utf8]{inputenc} 
\usepackage[T1]{fontenc}    
\usepackage{hyperref}       
\usepackage{url}            
\usepackage{booktabs}       
\usepackage{amsfonts}       
\usepackage{nicefrac}       
\usepackage{microtype}      
\usepackage{lipsum}
\usepackage{fancyhdr}       
\usepackage{graphicx}       
\graphicspath{{media/}}     
\usepackage{amsmath} 
\title{Urban Echoes: Decoding Transit Riders' Sentiments on Social Media for Smarter Mobility}

\author{
    Adway Das \\
    Department of Civil and Environmental Engineering   \\
    The Pennsylvania State University   \\
    State College, PA, USA 16802  \\
    \texttt{amd7293@psu.edu} \\
    \And
    Abhishek Kumar Prajapati \\
    Optym Inc. \\
    Dallas, Texas\\
    \texttt{prajapatipsu@gmail.com} \\
    \AND
    Pengxiang Zhang \\
    Department of Civil and Environmental Engineering \\
    The Pennsylvania State University \\
    State College, PA, USA 16802 \\
    \texttt{pengxiang@psu.edu} \\
    \And
    Mukund Srinath \\
    College of Information Sciences and Technology \\
    The Pennsylvania State University \\
    State College, PA, USA 16802 \\
    \texttt{mus824@psu.edu} \\
    \And
    Andisheh Ranjbari \\
    Department of Civil and Environmental Engineering \\
    The Pennsylvania State University \\
    State College, PA, USA 16802 \\
    \texttt{ranjbari@psu.edu} \\
}

\begin{document}
\maketitle

\begin{abstract}
Traditional transit surveys are resource-intensive and time-consuming, limiting their ability to address location-specific issues effectively. This study introduces an NLP-based framework that leverages real-time data from Twitter (currently X) as a pre-screening tool to refine and target transit agency surveys. The framework employs a two-step approach: Few-Shot learning classifies tweets into categories such as safety, reliability, and maintenance, while a lexicon-based sentiment analysis model evaluates sentiment polarity (positive, negative, neutral) and intensity. Additionally, spatial analysis maps sentiment trends to specific geographic areas, enabling transit agencies to pinpoint and prioritize problem zones. Applied to NYC subway tweets as a case study, the framework effectively identified service-related issues and localized concerns. A comparison with agency-run surveys confirmed its ability to uncover spatially targeted pain points, allowing agencies to design tailored surveys and prioritize improvements. This approach enhances feedback collection, addressing user concerns with greater precision and efficiency.
\end{abstract}

\keywords{Survey, Natural Language Processing (NLP), Sentiment Analysis, Transit, NYC Subway, twitter, social media, classification, Few-Shot Learning, Large Language Model (LLM).}

\section{Introduction}\label{sec1}
Public transit systems are integral to the vibrancy and sustainability of modern urban cities. They reduce traffic congestion, lower environmental pollution, and promote equitable access to essential services and opportunities. Efficient public transportation systems serve as the backbone of urban mobility, contributing significantly to the economic vitality and environmental health of cities. As urban populations continue to grow, the importance of reliable and accessible public transit becomes even more critical for sustainability and quality of life. Understanding user experiences within these complex systems is paramount for transportation authorities and policymakers aiming to enhance service quality and meet the evolving needs of commuters. Traditionally, user feedback surveys have been a helpful and effective mechanism for transit agencies to understand the needs and perspectives of their users. However, surveys can be costly, time-consuming, and limited in geographic coverage. On one hand, designing a well-structured survey that caters to the diverse population of a city is a daunting and expensive task. On the other hand, attempting to create a generalized survey to cover a large area with diverse populations and different needs may yield no helpful information while using precious limited resources of a transit agency. Research has shown that the median cost of conducting a survey per person for a transit agency is about \$36, with an average total cost of approximately \$350,000 for a medium-sized survey \cite{1Zalewski2019PublicTransitOD}.

Given the widespread use of social networks as communication channels, data from social media platforms offers a unique opportunity to gain real-time insights into the sentiments and experiences of users on a large scale. X (formerly known as Twitter), being one of the most popular social media platforms globally, hosts a massive volume of user-generated content daily. It has over 330 million active users generating approximately 500 million tweets per day, and projections indicate that these numbers will continue to grow in the coming years \cite{2Pfeffer2023JustAnotherDayTwitter}. Analyzing this data can provide valuable insights into public sentiment, preferences, and concerns related to various services, including public transportation. Natural language processing (NLP) has become a critical tool for analyzing large volumes of text-based data, particularly for sentiment analysis, enabling researchers to extract meaningful patterns from unstructured content. By harnessing advanced NLP techniques, transportation authorities and researchers can tap into this wealth of information to identify pain points, assess the effectiveness of implemented strategies, and make data-driven decisions. Importantly, social media data can reveal location-specific issues and sentiments, allowing transit agencies to recognize the distinct needs and challenges of different communities. This geographical granularity enables agencies to tailor surveys and interventions more effectively, focusing resources on specific areas where improvements are most needed. Even when not as detailed or precise as desired, the social media data can still provide general insights into certain issues and help reduce survey costs by narrowing down the thematic or geographical focus of future surveys.
However, to continuously monitor public opinion and utilize it for decision-making, it is essential to rely on user-generated data and develop systems capable of analyzing it efficiently. As a result, there has been a growing interest in employing text-based sentiment analysis and topic classification techniques to understand public opinions on a large scale \cite{3Modi2024SentimentTwitterFlask,4Khalid2022SentimentVaccinationTwitter,5Mahmud2022SentimentRideSharingApps},. Previous research has explored sentiment analysis and machine learning methods to categorize topics within transit-related tweets \cite{6Styawati2022Word2VecSVMTransportReviews,7Habib2023ImpactsCovidTwitterTransport}. Yet, a significant challenge remains: the need to manually or accurately label vast quantities of tweets, a crucial step for detailed classification tasks. This level of granularity is critical for transit agencies seeking to pinpoint and prioritize specific areas that require improvements in service quality. Our proposed framework addresses this challenge by enabling more precise and streamlined classification, allowing for targeted improvements based on user feedback.

This paper proposes an Artificial Intelligence (AI) based framework designed for transit agencies and city planners to address current operational challenges and optimize future service planning by leveraging insights from Twitter posts. The framework adopts a multi-faceted approach to gain a deeper understanding of public sentiment around specific transit-related issues. It deploys a three-step process: a) classifying tweets into distinct service-related categories (e.g., maintenance, schedule, safety, and security), b) analyzing the sentiment of these tweets by determining their positive or negative polarity, and c) understanding public ratings on particular issues while pinpointing high-concern locations for targeted intervention. To achieve these goals, we implemented VADER (Valence Aware Dictionary and sEntiment Reasoner), a well-established sentiment analysis tool known for its strong performance on social media data, to measure sentiment intensity in the tweet texts. Additionally, we employed Few-Shot learning, a cutting-edge topic classification technique that leverages pre-trained language models for classifying texts with minimal labeled data, reducing the need for large-scale manual annotation. By combining these methods, our framework efficiently processes large datasets of social media data or other types of text, while maintaining high accuracy in sentiment and topic classification. As a case study, we focused on the New York City (NYC) subway system, analyzing over 36,000 relevant tweets collected via Twitter's streaming API during 2022. These tweets were categorized into topics such as safety and security, scheduling, and maintenance issues, reflecting key areas of concern for NYC subway riders. Furthermore, our approach identifies specific geographic locations where recurring complaints or concerns are concentrated, enabling the Metropolitan Transportation Authority (MTA, the transit agency operating the NYC subway) to prioritize resource allocation for implementing targeted improvements. This data-driven framework offers an efficient solution to help transit agencies like the MTA make informed decisions, improving the daily experience of millions of transit riders. The remainder of this paper is organized as follows: Section \ref{sec2} presents the literature review, detailing prior studies on sentiment analysis and its application in transit planning. Section \ref{sec3} describes the data collection and methodology, including tweet classification and sentiment analysis approaches. Section \ref{sec4} discusses the results and insights derived from the NYC subway case study, and Section \ref{sec5} concludes with key findings and recommendations for transit agencies.

\section{Literature Review}\label{sec2}

\subsection{Sentiment Analysis in Public Transportation}

Sentiment analysis, also referred to as opinion mining or opinion analysis, involves determining the emotional tone of textual data by assessing its polarity (positive, negative, or neutral) and the intensity of the sentiment. In the domain of transportation, sentiment polarity is widely considered a categorical measure, with feedback typically classified as positive or negative \cite{8Peng2018CrossDomainSentimentACL,9Wang2014EnsembleLearningSentiment,10Xia2015WordPolarityBayesian}, positive, negative, or neutral \cite{11Mendez2019TwitterSatisfactionTransport,12Maghrebi2016TravelModeExtractionSocial,13Candelieri2015DetectingEventsTwitterUrban}, binary (zero or one) \cite{14OsorioArjona2021MadridMetroPerceptions,15Ali2017FuzzyOntologyTransportReviews}, or using a range such as one to five \cite{16Chua2022ChallengesMiningTwitterMalaysia} or minus one to plus one \cite{17Qi2020FrameworkTwitterOpinionsTransport}. Sentiment intensity, on the other hand, is often treated as a continuous measure to capture the strength of user opinions more precisely \cite{18Rane2018TwitterUSAirlineCOMPSAC,19Abiola2023SentimentCOVIDNigeriaVADER}. The granularity of sentiment analysis depends on the length and nature of the text being analyzed. Text-based sentiment analysis has been investigated on several levels: document level, sentence level, and phrase level. “Document-level sentiment analysis” is performed on entire documents, assigning a single polarity to the whole text. This method can be used to classify chapters or pages of a report as positive, negative, or neutral, using both supervised and unsupervised learning approaches \cite{20Bhatia2015BetterDocLevelSentimentRST,21Sharma2012DocLevelANNLexicons,22Tang2015SentimentSpecificRepresentation}. “Sentence-level analysis” examines each sentence individually, assigning separate polarity and intensity scores to each, which is useful for texts that cover a wide range of sentiments \cite{23Yang2014ContextAwareSentencePosterior,24Jagtap2013SentenceLevelApproachesSurvey,25Tackstrom2011SemiSupervisedLatentSentence}. The sentiment of individual sentences can be aggregated to represent the sentiment of the entire document or analyzed independently \cite{26Rao2018LSTMDocumentLevelSentiment}. Finally, “phrase-level analysis” focuses on individual words or phrases, calculating their aggregated sentiment and intensity. Social media data, given its brevity, is typically analyzed at the phrase level \cite{27Flek2020ReturningNtoNLP,28Zhang2014ExplicitFactorExplainableRec}.

In the literature, sentiment classification methods are generally divided into two categories: “machine learning-based methods” and “lexicon-based methods”. Machine learning approaches require a training dataset to learn from and a testing dataset to validate the model, while lexicon-based methods depend on predefined dictionaries of words and their associated polarity and intensity scores. Popular machine learning methods include Support Vector Machines (SVM), Naïve Bayes, and Decision Trees \cite{18Rane2018TwitterUSAirlineCOMPSAC, 24Jagtap2013SentenceLevelApproachesSurvey}, while more recent methods have incorporated advanced models like Google's BERT \cite{29Devlin2018BERTPretraining}, which uses contextual embeddings for sentiment classification. Lexicon-based methods rely on dictionaries such as SentiWordNet \cite{19Abiola2023SentimentCOVIDNigeriaVADER, 30Kucher2018VisualAnalysisSentimentStance}, Valence Aware Dictionary and sEntiment Reasoner (VADER) \cite{31Hutto2014VADER}, TextBlob \cite{32Santos2018SpatiotemporalTrafficSocial}, Afinn \cite{33Chaturvedi2019GeoSpatialSentimentUrbanTransport}, and Linguistic Inquiry and Word Count (LIWC) \cite{14OsorioArjona2021MadridMetroPerceptions,15Ali2017FuzzyOntologyTransportReviews}. While machine learning methods have shown promise, they tend to be data-intensive and require large amounts of labeled data, which can be difficult to obtain. Conversely, lexicon-based approaches are simpler, more interpretable, and still widely used because of their explainability. However, they may struggle with handling complex sentiment such as sarcasm or idiomatic expressions. Among lexicon-based methods, Afinn contains a dictionary of 3,300 words, each with an associated sentiment score, and has been one of the simplest and most popular lexicons for sentiment analysis \cite{30Kucher2018VisualAnalysisSentimentStance}. However, studies have shown that Afinn and older methods like SentiWordNet are less effective in handling nuanced language, especially in informal text like social media posts. More recent methods such as VADER and TextBlob have proven more effective. VADER, in particular, has gained popularity for analyzing user-generated text due to its ability to handle social media language, which is often informal and less structured than traditional textual data \cite{34Botchway2020BankTweetsSentiment,35Bonta2019LexiconBasedSurvey,36Elbagir2019TwitterSentimentNLTVADER}. It assigns both polarity and intensity scores, making it highly useful for classifying and quantifying the emotional content of social media posts. Sentiment analysis has been widely applied in the transportation sector to evaluate passenger satisfaction and identify service issues. For instance, a study on the Chicago Transit Authority used sentiment analysis on social media data to track real-time passenger sentiment during service disruptions, uncovering dissatisfaction patterns that were tied to specific operational issues \cite{37Haghighi2018TwitterTransitQoS}. Similarly, research on the London Underground categorized feedback into safety and punctuality concerns using both machine learning and lexicon-based sentiment analysis methods \cite{38Ulloa2016OpenInnovationTransportSM}. These studies demonstrate how sentiment analysis can be used to uncover patterns in public opinion and inform decision-making in urban transit systems. While sentiment analysis is an effective tool for extracting insights from text data, it faces several limitations. For instance, informal language, slang, and sarcasm prevalent in social media posts can skew sentiment results, making them less accurate. Additionally, machine learning methods, though promising, often require large amounts of labeled data, which may not always be available. Lexicon-based methods, while simpler and more interpretable, may fail to capture the subtleties of informal or complex language. Furthermore, domain-specific language used in transit-related feedback, such as transportation jargon or specific complaints, may require tailored lexicons to improve accuracy

\subsection{Use of Social Media Data and NLP in Transit and Urban Studies}
Social media platforms like Twitter provide a rich source of real-time, user-generated data that can be leveraged to analyze public sentiment toward transit services. By employing NLP techniques, transit agencies can extract meaningful insights from vast amounts of unstructured data. Sentiment analysis and topic classification are two of the most common NLP tasks applied in transit-related research. Social media sentiment analysis has been extensively used in urban studies to understand public opinion on transit issues. A notable study focused on New York City presents T-MAPS, a spatiotemporal model that leverages Twitter data to improve traffic insights by mapping tweets to regions and assigning traffic conditions using a time-varying digraph, with sentiment analysis and TF-IDF further enriching route descriptions \cite{32Santos2018SpatiotemporalTrafficSocial}. Similarly, a study on Singapore’s public transportation system employed Twitter sentiment analysis to track real-time changes in public opinion during peak hours \cite{39Hoang2016MicroEventTweetsTransit}. Both studies demonstrated how NLP can help uncover hidden patterns in user complaints and assist in real-time decision-making. The primary challenge of using social media data lies in its noise and variability. Posts are often short, informal, and lack the structured nature of traditional surveys. NLP models also face challenges in processing multilingual data, detecting sarcasm, and accounting for the bias inherent in social media demographics, as younger users tend to be overrepresented \cite{40Kinra2020TextualBigDataPolicy,41Hovy2021FiveSourcesBiasNLP}. Moreover, data privacy and ethical considerations must be managed carefully when analyzing publicly available social media data.

\subsection{Topic Classification and Challenges in Labeling and Classifying Large Datasets}
Topic classification in sentiment analysis allows researchers to categorize feedback into predefined themes such as safety, scheduling, or maintenance, which is crucial for providing actionable insights. Traditional classification methods like Naïve Bayes (NB), Support Vector Machines (SVM), and Random Forest (RF) have been widely used for text classification \cite{42Xu2017BayesianMultinomialNB,43Jiang2016DeepFeatureWeightingNB,44Maron1961AutomaticIndexing,45Shi2011ImprovedKNNDensity,46Shah2020ComparativeLRRFKNNText,47Chen2022AutomatedLegalTextRFDeep}. However, these methods often lack generalizability, especially when confronted with unseen data or domain-specific language like transit-related slang. A study in Toronto used topic classification to categorize social media posts related to the city's bus system, identifying delays and overcrowding as the most pressing concerns \cite{48Hosseini2018SocioSemanticTransitDebates}. In another study, a probabilistic topic modeling approach using Latent Dirichlet Allocation (LDA) was applied to transit smart card and land-use data to infer diverse activity patterns, enabling transportation authorities to identify and target specific non-commuting activities and understand behavioral shifts, such as those caused by COVID-19, for improved service planning \cite{49Aminpour2025MobilityPatternsSmartcardTopic}. One of the main challenges in topic classification is the reliance on extensive labeled datasets, which are often costly and time-consuming to create. Pre-trained models like Google’s BERT have made significant strides in improving classification accuracy, but these models still demand substantial computational power and large datasets to perform optimally. Moreover, many models struggle with generalizing to new, unseen data or non-standard language, such as the informal and dynamic language often found in social media \cite{50Qiu2020PretrainedModelsSurvey}. To overcome the issue of limited labeled data, pre-trained language models have emerged as valuable tools for text classification \cite{50Qiu2020PretrainedModelsSurvey,51Chronopoulou2019EmbarrassinglySimpleTransfer,52Radford2018ImprovingLanguageUnderstandingGPT}. These models are trained on vast amounts of text, enabling them to learn global semantic representations and significantly enhance performance across various NLP tasks, including text classification. For instance, large language models (LLMs) such as OpenAI’s GPT and LLaMA are pre-trained on trillions of tokens, focusing on next-word prediction. After this pre-training, they are fine-tuned on a wide range of tasks such as text classification, question answering, and text generation, improving their versatility and adaptability across different domains \cite{53Ouyang2022InstructHF}. An important innovation in these models is the incorporation of reinforcement learning based on human feedback, which allows them to follow task-specific instructions accurately, even for tasks they were not explicitly trained for. This capability enhances their generalization performance, making them highly adaptable for a variety of applications. One particularly powerful feature of LLMs is their ability to excel in Few-Shot Learning, where the model is provided with only a few examples of a task. By reviewing these examples, the model can infer the necessary patterns and rules to complete the task effectively, even without extensive task-specific training \cite{54Brown2020GPT3FewShot,55HosseiniAsl2024FewShotABSA,56Lin2022FewShotMultilingualGLMs}. This approach significantly enhances the model’s ability to deliver accurate results with limited labeled data, addressing one of the most critical limitations in topic classification: data scarcity. Few-shot learning is especially useful for handling dynamic, domain-specific, or evolving language, making it a valuable technique for analyzing user-generated content such as social media posts. For instance, in studies involving sentiment classification of movie reviews and product feedback, models trained using few-shot learning achieved comparable performance to fully supervised models, despite having access to only a limited number of labeled examples. In another application, intent classification within dialogue systems, few-shot learning techniques proved effective even in highly dynamic, domain-specific contexts, where data availability is often restricted. By reducing the dependence on large datasets and enabling models to generalize effectively across tasks, few-shot learning, in combination with pre-trained language models, represents a significant step forward in text classification. This approach offers a scalable and efficient solution in domains where labeled data are scarce or difficult to obtain, such as user-generated content on social media. The ability of few-shot learning to adapt to evolving and domain-specific language makes it an essential tool for modern text analysis, particularly in real-time contexts like public transportation feedback or sentiment analysis of transit services.

\subsection{Geospatial Analysis of Transit Sentiment}
Integrating sentiment analysis with geospatial data provides transit agencies with critical tools to pinpoint location-specific issues, enabling targeted service improvements where they are most needed. By linking public sentiment data with geographic insights, planners can identify areas with concentrated feedback, guiding interventions in high-impact locations within complex urban networks. For example, a study analyzing social media sentiment assessed public opinion on urban services, revealing clusters of recurrent complaints and prompting improvements in specific regions \cite{57Qi2019WirelessServicesGeospatialTwitter}. Similarly, a case study in Sydney’s transit system used Twitter sentiment to map passenger experiences, highlighting concerns such as delays and overcrowding on various lines, which informed data-driven adjustments \cite{58Lock2020PassiveGeoParticipationTransport}. These examples illustrate how geospatial sentiment analysis allows urban planners to respond to localized feedback, optimizing resources and addressing the most pressing service concerns effectively. However, geospatial analysis faces several challenges, including data availability and precision. Social media posts often lack explicit geographic tags, and determining the location from the content of a tweet can be inaccurate. Furthermore, while geospatial sentiment analysis can provide high-level insights, it often struggles with the granularity required for specific operational improvements. Bias in the data, where users in certain regions are more likely to post feedback, also skews the results \cite{59KovacsGyori2020GeoAnalysisUrbanLivability}.

Despite the widespread use of sentiment analysis in transit studies, there remains a significant research gap in incorporating spatial analysis and topic classification using social media data across various transit-related topics. This study is motivated by the need to address this gap, offering location-specific insights into different concerns expressed by users, ultimately enhancing the understanding of public sentiment and helping transit agencies make more informed, data-driven decisions.

\section{Data Collection}\label{sec3}
\begin{figure}
    \centering
    \includegraphics[width=0.8\linewidth]{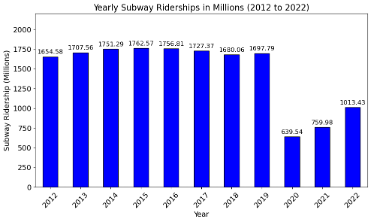}
    \caption{MTA Subway Annual Ridership Numbers From 2012 To 2022}
    \label{fig1}
\end{figure}

The New York City (NYC) subway system is an extensive and intricate transportation network, spanning over 800 miles of tracks throughout the bustling NYC metropolitan area. Historically, its annual ridership has been around 1.7 billion from 2012 to 2019, as shown in Figure \ref{fig1}. However, due to the COVID-19 outbreak, the annual ridership drastically declined to less than 0.64 billion. Despite this reduction in ridership, a significant amount of data continues to be generated by the subway users, making it an invaluable source for sentiment analysis. To showcase the effectiveness of the proposed framework, the NYC subway system was selected as the region of interest. By focusing on this specific transit network, the framework aims to analyze the vast amount of data available through social media platforms, particularly Twitter.

\begin{figure}
    \centering
    \includegraphics[width=0.8\linewidth]{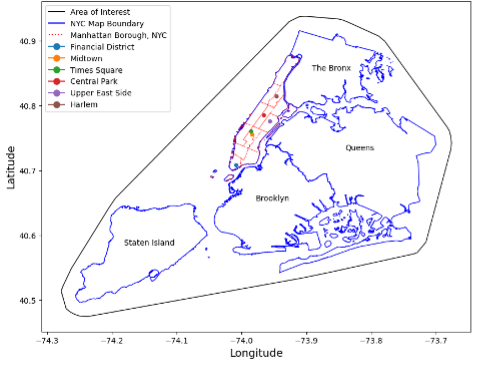}
    \caption{Illustration of The Area of Interest for Data Collection and Some Critical Locations of Manhattan Borough, NYC}
    \label{fig2}
\end{figure}

Tweets related to the NYC subway system were collected through the Twitter API and using a web scraping technique, specifically using 'snscrape,' a free scraper tool designed for social networking services like Twitter. The data collection process involved specifying relevant search keywords, hashtags, and date ranges to extract tweets pertinent to the NYC subway system. To ensure comprehensive data collection, a combination of 10 unique search words, including "MTA," "NYC SUBWAY," "Subway Station," "New York subway," and "Subway transit," along with 12 relevant locations within the MTA subway network, was utilized.
To ensure a comprehensive collection of tweets related to the NYC subway system, we have considered a slightly larger region than the subway service region. By expanding the data collection area, we aim to capture a wider range of tweets that pertain to the subway system and its associated topics. Figure \ref{fig2} illustrates the expanded region selected to gather pertinent Twitter data, along with several critical locations within the Manhattan Borough.

\begin{figure}
    \centering
    \includegraphics[width=0.8\linewidth]{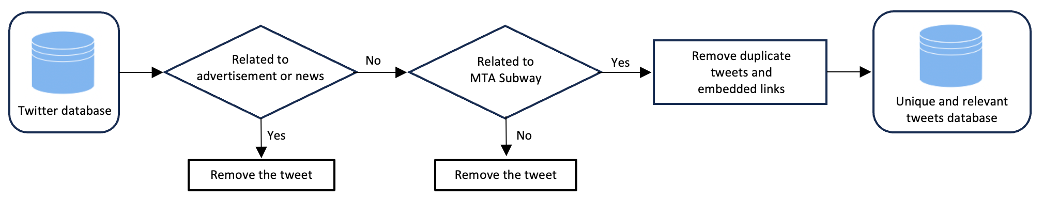}
    \caption{The Filtering and Processing Steps Adopted for Data Collection.}
    \label{fig3}
\end{figure}

After extracting the tweets, a filtering process was implemented to remove duplicate tweets from the same user and to eliminate any embedded links within the text. The goal of this filtering step was to ensure that the dataset consisted of unique and relevant tweets only. Figure \ref{fig3} shows an overview of the data collection steps utilized. The sentiment analysis was carried out using a model that does not require any preprocessing steps, such as the removal of stop words or punctuations from the input data. Therefore, the stop words and punctuations present in the tweets were not removed. 

By utilizing this approach, a rich dataset of 102,530 tweets specifically related to the NYC subway system that were posted during 2022 was extracted. The dataset included the text of the tweets, the time and date of posting, and the geo-tagged location associated with the tweets. Out of this pool of tweets, we randomly sampled 36,000 tweets to use for the analysis.

\section{Methodology}\label{sec4}

This study presents a novel NLP-based framework for analyzing Twitter data to gain insights into user experiences in a transit system. By leveraging inexpensive Twitter data, this framework could serve as a general screening tool, assisting researchers and decision-makers in conducting user feedback surveys more efficiently and at a reduced cost compared to traditional methods, by focusing on high-concern topics and/or regions. 

The core of this proposed framework revolves around two crucial models: a sentiment analysis model and a tweet classification model. The sentiment analysis model, known as VADER, calculates sentiment scores for each tweet, providing a deep understanding of users’ emotions and perceptions regarding the transit system. For tweet classification, we use OpenAI’s GPT-3.5 Turbo, a large language model that has been shown to effectively perform classifications when provided with a task definition. We use GPT-3.5 to categorize the tweets into predefined topics related to user experience on the transit system, allowing for streamlined analysis and organization of user feedback. By seamlessly integrating these models, the framework can efficiently process and analyze vast volumes of Twitter data to extract valuable insights into user sentiments about specific transit-related topics.

\subsection{Classification of Tweets Using Few-Shot Learning}

While users share short texts on various topics on Twitter, obtaining labeled data suitable for training machine learning models can be extremely challenging. Labeling data involves the manual process of assigning specific categories or classes to the tweets, which requires human effort and expertise. However, due to the large volume of tweets and the diversity of topics discussed on the platform, it becomes impractical and costly to manually label a sufficient number of tweets to train effective models for various Natural Language Processing (NLP) tasks, such as text classification. To tackle the text classification issue effectively, we deployed a few-shot learning approach.

Few-shot learning is a machine learning approach to that specifically addresses the challenges of training models using limited labeled data \cite{54Brown2020GPT3FewShot}. In the realm of NLP-based tasks like text classification, traditional supervised learning methods often demand a vast amount of labeled data to achieve good generalization and accuracy. Conversely, few-shot learners adopt a different approach. They rely on a pre-trained language model, such as GPT-3.5 Turbo, and need only a small number of labeled samples to prime the model to solve tasks similar to the provided labeled samples. The power of pre-trained language models like GPT-3.5 Turbo lies in their extensive pre-training on massive datasets containing billions of sentences from diverse sources. This pre-training endows the model with a broad understanding of language, syntax, semantics, and the intricate connections between words and phrases. The models are subsequently finetuned on a large number of varied tasks, such as text classification, question answering, textual entailment, coreference resolution and text generation. Finally, they are trained using reinforcement learning to follow task instructions based on human feedback. This training procedure allows models to accurately follow instructions and even solve tasks on which they were not previously trained. These models perform especially well when in addition to task instructions, examples are also provided, which is the case for few-shot learning. As a result, these types of models are able to effectively generalize based on provided prompts and task definitions to solve a wide range of problems \cite{52Radford2018ImprovingLanguageUnderstandingGPT,55HosseiniAsl2024FewShotABSA,56Lin2022FewShotMultilingualGLMs} . In addition, when models are prompted with a few labeled samples, they can quickly adapt to the specific characteristics of the target task, further improving the performance. The model’s adaptability and generalization capabilities have shown promising results, even when trained with only limited labeled samples \cite{53Ouyang2022InstructHF,60Kamal2021HostilityDetectionHindi,61Gunel2021SupContrastFinetuning}. Using a pre-trained language model for text classification significantly reduces the data annotation burden, as only a small subset of data needs to be labeled. Also, due to the vast pre-training, the model can leverage the knowledge learned from diverse contexts to better generalize and make accurate predictions on new, unlabeled instances. 
In this study, we used a qualitative coding approach to create tags for different categories of user experiences on the transit system. We sampled 500 tweets and assigned each Tweet a category based on the aspects of user experience described in the Tweet. We merged similar categories and ended up with three categories that exhaustively describe user experiences of the transit system and a fourth category that did not convey any such personal user experience. The four categories are as follows.

\begin{itemize}
    \item \textbf{Cleanliness and Maintenance}: This category encompasses tweets discussing issues on the cleanliness and maintenance of the subway system.
    \item \textbf{Schedule and Operations}: This category includes tweets that specifically address the schedule of subway trains, including delays, timeliness, and related topics.
    \item \textbf{Safety and Security}: tweets falling into this category highlight concerns, incidents, or discussions related to the safety and security of subway users.
    \item \textbf{Other}: This category captures Tweets unrelated to user experience on the transit system. 
\end{itemize}

It is worth mentioning that the 'Other' category is crucial for identifying tweets that, while expressing strong emotional content, do not relate directly to the user experience on the transit system. This includes a broad range of subjects from personal anecdotes to broader social or political commentary. By effectively isolating these tweets, our framework enhances the accuracy and fairness of classifying the other three categories. This separation ensures that the sentiments and opinions relevant to the other three categories are not overshadowed or diluted by unrelated content, leading to more precise insights for transit authorities. 

To perform the classification task, we conducted experiments to tune the GPT-3.5 model for tweet classification using a varying number of tweets, ranging from 1 to 10. We found that the model's performance showed diminishing returns beyond five tweet samples per category. As a result, we decided to use five-tweet samples from each topic category considered in this study. This struck a good balance between performance and resource efficiency.

\subsubsection{Evaluation metrics for tweet classification}

In the context of tweet classification, True Negatives (TN), False Negatives (FN), True Positives (TP), and False Positives (FP) are used to describe the results of predictions made by the classification model. It should be noted that in this context, "positive" and "negative" refer to the class labels predicted by the model, representing the classes that the model identifies. These differ from "positive sentiment" and "negative sentiment," which refer to the emotional tone or polarity of the tweets. The four labels for our analysis are described in the following.

\begin{itemize}
    \item TN refers to the number of tweets that are correctly classified as "Other" by the model. These are the tweets that do not belong to any of the specific categories (Service and Maintenance, Schedule, or Safety and Security), and the model accurately identifies them as such. The range of TN is from 0 to the total number of tweets that do not belong to any of the targeted categories. 
    \item FN represents the number of tweets that are misclassified as "Other" by the model while they belong to one of the three positive categories. These tweets are falsely rejected or missed by the model, and it fails to identify their actual category. The range of FN is from 0 to the total number of tweets belonging to the positive categories. 
    \item TP denotes the number of tweets that are correctly classified into one of the three positive categories. The model accurately identifies these tweets as belonging to the respective positive category. The range of TP is from 0 to the total number of tweets belonging to the positive categories. 
    \item FP refers to the number of tweets that are incorrectly classified into one of the positive categories by the model, while they belong to the "Other" category. These tweets are falsely identified as belonging to a positive category. The range of FP is from 0 to the total number of tweets that are in the "Others" category.
\end{itemize}

TN, FN, TP, and FP values for each category are used to evaluate the performance of the classification model by calculating metrics like Precision, Recall, and F1 score \cite{62Goutte2005ProbabilisticInterpretationFscore}. Each of the metrics is described in detail next.

\begin{equation}
\text{Precision} = \frac{TP}{TP + FP}
\label{eq1}
\end{equation}

Recall, also known as Sensitivity or TP rate, is calculated by dividing the number of TP by the sum of TP and FN, as shown in Equation 2. It represents the proportion of correctly identified positive instances among all instances that belong to the positive class. Recall measures how many of the actual positive instances were correctly identified by the model, indicating the model's ability to capture all relevant positive instances.

Precision is calculated by dividing the number of TP by the sum of TP and FP, as shown in Equation 1. It measures the proportion of correctly identified positive instances among all instances that the model classified as positive. In other words, precision indicates how many of the positive predictions made by the model are correct. A high precision score suggests that the model has a low false positive rate and is accurate in identifying positive instances. 

\begin{equation}
\text{Recall} = \frac{TP}{TP + FN}
\label{eq2}
\end{equation}

The F1 Score is the harmonic mean of precision and recall and provides a balanced measure that considers both metrics. It is calculated by dividing twice the product of precision and recall by their sum, with the formula represented in Equation 3. The F1 Score ranges from 0 to 1, where a higher score indicates better model performance. It is particularly useful when dealing with imbalanced class distributions, as it considers both false positives and false negatives in determining the overall effectiveness of the model.

\begin{equation}
\text{F1 Score} = \frac{2 \cdot (\text{Precision} \cdot \text{Recall})}{\text{Precision} + \text{Recall}}
\label{eq3}
\end{equation}

\subsection{Sentiment Analysis Using VADER}
\begin{figure}
    \centering
    \includegraphics[width=0.8\linewidth]{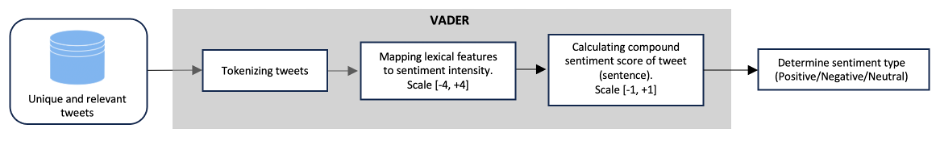}
    \caption{The Process of VADER Sentiment Analysis}
    \label{fig4}
\end{figure}

In NLP-based sentiment analysis, selecting an appropriate tool or package is pivotal to the accuracy and effectiveness of the analysis. The Valence Aware Dictionary and sEntiment Reasoner (VADER)\cite{36Elbagir2019TwitterSentimentNLTVADER} is a widely used lexicon-based sentiment analysis method that has been specifically tailored to capture sentiments prevalent in social media texts. VADER exhibits responsiveness to both polarity (positive/negative) and intensity (strength) of emotions, making it highly suitable for analyzing text-based sentiments \cite{63Park2018SentimentAIICIEA}. It is a built-in component of the widely used Natural Language Toolkit (NLTK) and can be readily applied to unlabeled/balanced text data. The underlying principle of VADER sentiment analysis lies in its reliance on a pre-built dictionary of sentiment lexicons that maps lexical features to emotion intensities, represented as sentiment scores.  These sentiment lexicons were designed to contain sentiment intensity scores for a wide range of words, phrases, and emoticons based on human annotations. By summing up the intensity of each word in a given text, VADER generates a sentiment score that reflects the overall sentiment expressed within the text. For instance, words such as "perfect" and "nice" are typically associated with positive sentiments. Notably, VADER exhibits exceptional linguistic intelligence by recognizing the nuanced meanings of words, even in complex cases such as "was not nice," where the sentiment is negative. Furthermore, VADER considers capitalization and punctuation emphasis, thereby capturing the intensity of sentiments, e.g., it recognizes that "ENJOY!!!" expresses a higher level of positive intensity compared to "enjoy." It is worth noting that VADER does not require preprocessing steps such as removing stop words (e.g., "the", "and", "is", "in", etc.) or punctuations (e.g., periods, commas, exclamation marks) from the input data, leveraging both the content and context of the text to comprehend the overall sentiment of a sentence. An overview of the overall process of VADER sentiment analysis used in this study is depicted in Figure \ref{fig4}. In VADER, individual words of the phrases are used to map their lexical features to sentiment scores within the range of -4 to 4. To calculate sentiment at the sentence level, the scores of each word in the sentence are summed up and normalized using Hutto normalization (shown in Equation 4) to obtain a compound sentiment score ranging from -1 to +1. This normalization guarantees consistency and facilitates a more accurate understanding of the overall sentiment expressed in a sentence.

\begin{equation}
CSC_i = \frac{x_i}{\sqrt{x_i^2 + \alpha}}
\label{eq4}
\end{equation}

In this equation, $x_i$ is the sum of the sentiment scores of the constituent words of the tweet i, and $\alpha$ is the normalization parameter that was set to 15 as suggested by \cite{24Jagtap2013SentenceLevelApproachesSurvey} for short text like tweets. Using the calculated compound sentiment score for tweets, we classified tweets into three emotion levels (positive, neutral, and negative) based on the following thresholds. Tweets with compound sentiment scores greater than 0.1 were categorized as having a positive sentiment, while those with compound sentiment scores less than -0.1 were considered to convey negative sentiments. Tweets with compound sentiment scores between -0.1 and 0.1 were classified as neutral emotions. Please note that since VADER is a mathematical tool, there is no evaluation metric to measure its performance.

\section{Results and Discussions}\label{sec5}
\subsection{Validation of Tweet Classification Model}
\begin{figure}
    \centering
    \includegraphics[width=0.8\linewidth]{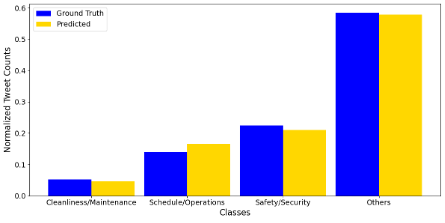}
    \caption{Comparison of Predicted and Ground Truth Labels for 500 Random Samples.}
    \label{fig5}
\end{figure}

Since tweets are unlabeled, validation of the classification prediction is crucial. To validate the tweet classification model, a random sample of 500 tweets was taken from the dataset used for classification. These 500 tweets were manually annotated by the authors with proper categories and were used as the ground truth. Figure \ref{fig5} presents the comparison between the ground truth and the predicted results for the sample of 500 tweets, demonstrating a good match between the two.

The validation metrics for the tweet classification model are also presented in Table 1. The high values of precision, recall, and F1 score metrics demonstrate the effectiveness of our approach in accurately categorizing the tweets. The high precision and recall scores indicate the model's ability to correctly identify tweets within each category, and the high F1 score further reinforces the reliability of our classification model.

\begin{table}[ht]
\centering
\caption{Validation metrics for the classification model}
\label{tab1}
\begin{tabular}{|c|c|c|c|}
\hline
\textbf{Metrics} & \textbf{Precision} & \textbf{Recall} & \textbf{F1 Score} \\
\hline
\textbf{Value} & 0.9456 & 0.9420 & 0.9425 \\
\hline
\end{tabular}
\end{table}

\subsection{Tweet Classification Model Results}
The analysis dataset was created by randomly sampling 36,000 tweets. In Table 2, the distribution of predicted categories for all the tweets is shown. According to the results, safety and security are identified as the highest concerns, constituting approximately 21.5\% of the tweets by the NYC subway users. They are followed by schedule-related issues, constituting about 16.8\% of the tweets by the NYC subway users. Valuable insights into the major areas of concern expressed by the public in relation to the NYC subway system can be gained from these findings.

\begin{table}[ht]
\centering
\caption{Number of tweets in each category after classification}
\label{tab2}
\begin{tabular}{|c|c|c|c|c|}
\hline
\textbf{Classes} & \textbf{Cleanliness/Maintenance} & \textbf{Schedule/Operations} & \textbf{Safety/Security} & \textbf{Others} \\
\hline
\textbf{Number of tweets} & 1667 & 6050 & 7708 & 20575 \\
\hline
\textbf{Category percentage (\%)} & 4.6 & 16.8 & 21.5 & 57.1 \\
\hline
\end{tabular}
\end{table}

\subsection{Sentiment Analysis Model Results}
The VADER method was employed to calculate the compound sentiment score of tweets in a database consisting of 36,000 tweets. Example results of this analysis are presented in Table 3, featuring a selection of 8 tweets from various categories, along with their corresponding sentiment scores and emotion levels. The first column of the table displays the actual content of each tweet, while the "Coordinates" column indicates the specific online locations where these tweets were originally posted. The "Category" column shows the tweet category predicted by the tweet classification model. Lastly, the "Sentiment" column represents the overall sentiment of each tweet related to its category. This provides strong evidence of the proposed framework’s ability to comprehend the intricate and sarcastic nature of tweets and accurately categorize and evaluate its sentiment.

\begin{table}[ht]
\small
\setlength{\tabcolsep}{4pt}
\renewcommand{\arraystretch}{1.2}
\centering
\caption{Examples of tweets in the analysis dataset, along with their identified category and calculated compound sentiment score}
\label{tab3}
\begin{tabular}{|p{6.3cm}|p{2.6cm}|p{2.5cm}|p{2.2cm}|p{1.8cm}|}
\hline
\textbf{Posted tweets} & \textbf{Coordinates} & \textbf{Category} & \textbf{Compound score} & \textbf{Sentiment} \\
\hline
Why would you WANT to ride the subway without a mask? It is so stinky ... so stinky. & 40.683935, -74.026675 & Cleanliness and maintenance & -0.6651 & Negative \\
\hline
Moynihan Train Hall is shaping up to be very nice just do not look underneath at the rest of Penn Station. & 40.683935, -74.026675 & Cleanliness and maintenance & 0.4754 & Positive \\
\hline
"HOW HOW HOW ARE TRAINS BAD ON A RANDOM THURSDAY BEFORE 9PM?" bc in NYC u’ll miss anything is possible. & 40.683935, -74.026675 & Schedule and operations & -0.7034 & Negative \\
\hline
@NYCTSubway whoever is the conductor that kept the doors open for me and a few other people on the 6 train just now (car 2040) is my hero. Saved me from a long wait and I need to get home. Thank you!!! & 40.683935, -74.026675 & Schedule and operations & 0.7701 & Positive \\
\hline
Paid \$2.75 to be in a subway car with a loud and aggressive man threatening to hit his female partner. Switched cars at next stop to be in a public toilet / urine-odor, crowded car for the rest of my ride. & 40.570842, -74.041878 & Safety and security & -0.7351 & Negative \\
\hline
Fascinating reporting @MolaReports after \#NYPD captured the Subway shooter in NYC. It's amazing all victims from yesterday's incident are okay thank goodness. & 40.495865, -74.255641 & Safety and security & 0.9517 & Positive \\
\hline
Spent this entire train smiling and playing with the cutest chubby cheek baby. She smiled the whole way. What a beautiful moment of calm and joy to cherish before I enter back into the depths of hell aka the west village. & 40.683935, -74.026675 & Others & 0.9337 & Positive \\
\hline
@RockawayRose @MTA You can ride the subway. Be alert! & 40.541722, -73.962582 & Safety and security & -0.0867 & Neutral \\
\hline
\end{tabular}
\end{table}

\begin{figure}
    \centering
    \includegraphics[width=0.8\linewidth]{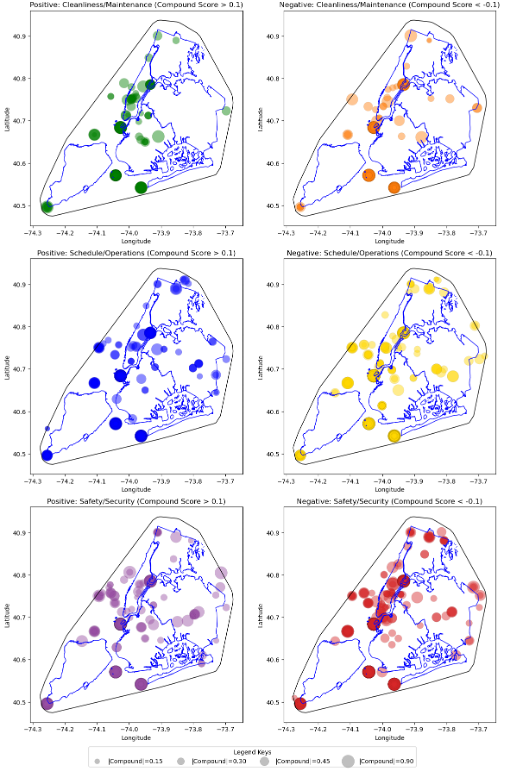}
    \caption{Spatial Distribution of Tweets by Different Categories}
    \label{fig6}
\end{figure}

Figure \ref{fig6} depicts the spatial distribution of positive and negative tweets, along with their identified categories and calculated sentiment scores. In this analysis, it was assumed that any user’s tweet about a specific issue shortly (within a few minutes) after observing it, so the tweets occurring within a 1-mile radius of a subway station were mapped to that station, utilizing the geo-tagged locations of the tweets. Each circle in the figure represents the aggregation of tweets that fall within a 1-mile distance from a subway station. The size of each circle indicates the cumulative absolute value of the compound sentiment within that 1-mile radius of the corresponding subway station, with larger circles signifying a higher overall magnitude of sentiment, whether positive or negative. The color intensity of each circle represents the number of tweets found within that 1-mile radius: darker shades indicate a higher volume of tweets, whereas lighter shades indicate fewer tweets in that specific location. This visualization method offers valuable insights into the distribution of positive and negative sentiments around subway stations, enabling the understanding of prevalent sentiments and trends in different areas.

\begin{figure}
    \centering
    \includegraphics[width=0.8\linewidth]{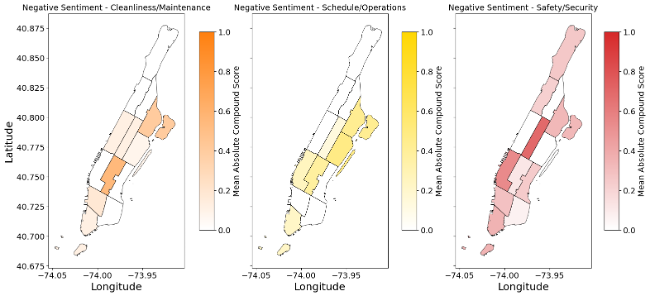}
    \caption{Key Areas of Concern in Manhattan Borough, NYC Related to Different Categories.}
    \label{fig7}
\end{figure}

This spatial distribution can help an agency like MTA identify specific locations with recurring concerns and consequently focus their efforts and resources there to implement targeted improvements. For example, the plots show that safety and security concerns are more concentrated in Midtown and Financial District, while issues about schedule are more pronounced in Upper Manhattan and Queens. This information can also help transit agencies narrow down the thematic and geographical focus of their future surveys by focusing only on stinging points, thus reducing the high cost of surveys.  
To further illustrate, Figure \ref{fig7} presents the distribution of mean compound sentiment scores of tweets across all districts in the Manhattan borough classified by tweet category. It offers insights into the sentiments expressed in tweets for three specific issue types: "Cleanliness/Maintenance," "Service/Schedule," and "Safety/Security." It is observed that areas like Times Square or Central Park consistently exhibit the highest number of negative tweets across all three categories. This result could be attributed to the area's high footfall due to tourism. Additionally, it is noteworthy that the Upper East Side and East Harlem show a significant concentration of negative tweets concerning safety and security issues. More information can be gained from conducting further investigation on the reasons behind this trend. This visualization of results highlights how the proposed framework can empower transit agencies and city planners to identify the high-concern areas for focusing their efforts and resources into implementing relevant improvements.

\begin{figure}
    \centering
    \includegraphics[width=0.8\linewidth]{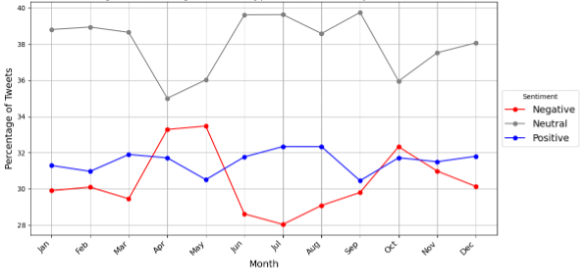}
    \caption{Temporal Distribution of User Sentiments for The Year 2022.}
    \label{fig8}
\end{figure}

The temporal distribution of sentiments and the understanding of how overall sentiment changes over time would also be found helpful by policymakers and transit agencies as it allows the assessment of the impact of implemented strategies or external factors. Figure \ref{fig8} presents the temporal distribution of positive, negative, and neutral sentiments for 2022, revealing that subway users were most satisfied with the system in March and during the summer months (June-September). This finding aligns with the observations from the Fall 2022 MTA customer survey, which reported that 54\% of subway customers expressed satisfaction with the service, indicating an increase of six percentage points from the Spring 2022 survey \cite{64MTA2023CustomersCountFall2022}. Specifically, the percentage of negative tweets in April and May of 2022 was around 33\%, and it was reduced to around 28\% in June-August 2022, depicting an increase in customer satisfaction. The MTA survey also highlighted that while concerns about safety, security, and service reliability persisted, satisfaction in these areas increased compared to the Spring 2022 survey. Notably, satisfaction with personal security increased by seven percentage points in stations and nine percentage points on trains, reflecting improved perceptions of safety. These congruent results between our sentiment analysis and the MTA's customer survey underscore the effectiveness of using social media data to gauge public perception and response to transit system improvements. This highlights how inexpensive social media data can be used to supplement and, in some cases, replace costly user surveys for transit agencies, providing timely and actionable insights.

By employing our proposed framework, a thorough analysis of user sentiments is conducted, encompassing both spatial and temporal dimensions. This approach allows for the identification of intricate patterns and trends, thereby providing valuable insights into the nuanced distribution of user perspectives. The knowledge acquired through these investigations can lead to cost savings for agencies and prove beneficial for policymakers and transportation authorities, as it facilitates data-informed decision-making and strategic implementation of targeted service enhancements.

\section{Conclusions}\label{sec6}
This study proposed a novel framework that empowers transit agencies and city planners to tap into the vast pool of inexpensive social media data, serving as a viable replacement or supplement to costly user feedback surveys. The primary objective of this research was to showcase the potential of utilizing widely accessible social media data to understand users' concerns regarding a particular issue as well as the issue's major occurrence locations, employing advanced Natural Language Processing (NLP) techniques. Specifically, the proposed framework adopts a combination of two powerful methods: VADER-based sentiment analysis and Few-Shot learning-based classification model. This synergistic approach enables the extraction of valuable insights from Twitter data, allowing a deeper understanding of users' sentiments related to specific issues. To effectively demonstrate the framework's capabilities, we focused on analyzing tweets pertaining to the NYC subway system. With a comprehensive collection and examination of 36,000 tweets, the study successfully discerned the underlying sentiments expressed in tweets, further categorizing them into three distinct thematic areas: \textit{Cleanliness and Maintenance, Schedule and Operations, Safety and Security.}

The analysis indisputably revealed safety and security as prominent issues of concern, originating from specific subway stations and neighborhoods. Such information empowers decision-makers to take swift action, allocating necessary resources and implementing targeted measures to address safety and security concerns in the highlighted locations. By doing so, agencies can potentially enhance the overall user experience, promote transportation equity, and positively influence the public perception of the transit system. Addressing these concerns in underserved or high-risk areas can help ensure that all communities, regardless of socioeconomic status, have access to safe and reliable transportation options, a key pillar of transportation equity. Similarly, the examination of tweets concerning schedule and cleanliness highlighted clusters of dissatisfaction surrounding particular subway lines or sections. With this powerful knowledge at their disposal, transit agencies can prioritize efforts to improve punctuality and maintenance in critical segments, especially in neighborhoods that may have been historically overlooked. By focusing on equitable resource distribution, transit systems can better serve all users, fostering inclusivity and fairness across the network. This proactive approach not only increases user satisfaction and instills greater confidence in the system's reliability and efficiency, but also contributes to the creation of a more sustainable and modern city, where all residents have access to high-quality transit services that meet the needs of the population equitably.

The versatility of the proposed framework extends far beyond the NYC region, transit systems and even the transportation industry, making it applicable in diverse regions, at various times, and across multiple service providers, including airports, buses, parking garages, and ride-share services. Its adaptability opens a wide array of possibilities for utilization in different contexts. One such application lies in transit fare policy, where decision-makers can effectively use the framework to assess how changes in fares impact user satisfaction, identifying the specific areas most affected by such modifications. The strength of this framework lies in its ability to gauge user feedback across various service contexts. By leveraging social media data and advanced NLP techniques, it provides valuable insights into users' perceptions and opinions related to specific issues. Whether it is identifying safety concerns, evaluating service quality, or soliciting customer feedback about the transit agency, this framework serves as a powerful tool for decision makers to make data-driven choices and ensures customer-centric improvements in a wide range of industries and service sectors. Additionally, this framework can be a powerful tool to identify spatial and the temporal distribution of critical problems in an area. This capability can be very helpful for conducting tailored surveys for specific locations or optimizing and hastening resource distribution to identify and mitigate the respective issues.

\bibliographystyle{unsrt}  
\bibliography{references}

\end{document}